\documentclass[journal,twoside,print]{ieeecolor}
\usepackage{tmi}
\usepackage{cite}
\usepackage{amsmath,amssymb,amsfonts}
\usepackage{algorithmic}
\usepackage{graphicx}
\usepackage{textcomp}
\usepackage{tabularx}

\begin{document}
\title{PADReg: Physics-Aware Deformable Registration Guided by Contact Force for Ultrasound Sequences}
\author{Yimeng Geng, Mingyang Zhao, Fan Xu, Guanglin Cao, Gaofeng Meng, \IEEEmembership{Senior Member, IEEE}, Hongbin Liu
\thanks{This work has been submitted to the IEEE for possible publication. Copyright may be transferred without notice, after which this version may no longer be accessible.}
\thanks{This work was supported in part by the InnoHK programme.}
\thanks{Yimeng Geng, Fan Xu, Guanglin Cao, Gaofeng Meng are with the State Key Laboratory of Multimodal Artificial Intelligence Systems, Institute of Automation, Chinese Academy of Sciences, Beijing, China and School of Artificial Intelligence, University of Chinese Academy of Sciences, Beijing, China (email: gengyimeng2022@ia.ac.cn, xufan181@mails.ucas.ac.cn, caoguanglin2021@ia.ac.cn, gaofeng.meng@ia.ac.cn  ).}
\thanks{Mingyang Zhao, Gaofeng Meng are with Center for Artificial Intelligence and Robotics, HK Institute of Science \& Innovation, Chinese Academy of Sciences, Hong Kong SAR (email: mingyang.zhao@cair-cas.org.hk, gaofeng.meng@ia.ac.cn)}
\thanks{Hongbin Liu is with the State Key Laboratory of Multimodal Artificial Intelligence Systems, Institute of Automation, Chinese Academy of Sciences, Beijing, China, Center for Artificial Intelligence and Robotics, HK Institute of Science \& Innovation, Chinese Academy of Sciences, Hong Kong SAR and School of Biomedical Engineering and Imaging Sciences, King’s College London, London, UK (email: liuhongbin@ia.ac.cn)}
}
\maketitle

\begin{abstract}
Ultrasound deformable registration estimates spatial transformations between pairs of deformed ultrasound images, which is crucial for capturing biomechanical properties and enhancing diagnostic accuracy in diseases such as thyroid nodules and breast cancer. 
However, ultrasound deformable registration remains highly challenging, especially under large deformation. The inherently low contrast, heavy noise and ambiguous tissue boundaries in ultrasound images severely hinder reliable feature extraction and correspondence matching. Existing methods often suffer from poor anatomical alignment and lack physical interpretability. 
To address the problem, we propose PADReg, a physics-aware deformable registration framework guided by contact force. PADReg leverages synchronized contact force measured by robotic ultrasound systems as a physical prior to constrain the registration. 
Specifically, instead of directly predicting deformation fields, we first construct a pixel-wise stiffness map utilizing the multi-modal information from contact force and ultrasound images. The stiffness map is then combined with force data to estimate a dense deformation field, through a lightweight physics-aware module inspired by Hooke's law. 
This design enables PADReg to achieve physically plausible registration with better anatomical alignment than previous methods relying solely on image similarity. Experiments on in-vivo datasets demonstrate that it attains a HD95 of 12.90, which is 21.34\% better than state-of-the-art methods. 
The source code is available at https://github.com/evelynskip/PADReg.
\end{abstract}

\begin{IEEEkeywords}
Deformable Registration, Force Fusion, Physics-aware Ultrasound Imaging, Robotic Ultrasound
\end{IEEEkeywords}

\section{Introduction}
\label{sec:introduction}
\begin{figure}[!t]
\centerline{\includegraphics[width=\columnwidth]{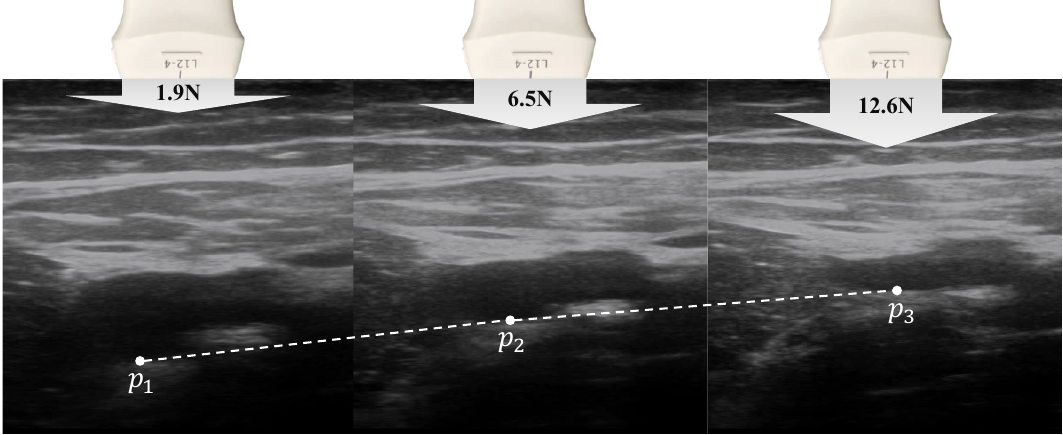}}
\caption{Illustration of pressure-induced deformation. Left, middle and right images are ultrasound images when transducer contact force increases. $p_1$, $p_2$ and $p_3$ are the same keypoint in different images. As the contact force increases, anatomical structures are compressed.}
\label{fig1}
\end{figure}

\IEEEPARstart{U}{ltrasound} imaging has been widely adopted in clinical practice due to its inherent advantages as a non-invasive, radiation-free, portable, and real-time modality. Unlike non-contact modalities such as computed tomography (CT) and magnetic resonance imaging (MRI), ultrasound requires transducer contact with the skin. To obtain high-quality ultrasound images, transducer contact force must be applied, which causes tissue deformation that is positively related to the applied force (Fig. \ref{fig1}). 

Ultrasound deformable registration, which aims to estimate spatial transformations between pairs of deformed ultrasound images \cite{sotiras2013deformable}, is vital for various clinical applications \cite{ricci2014clinical,cantisani2015strain,onur2015utility}. 
On the one hand, contact-induced deformation can impair the consistency of lesion tracking and measurement \cite{burcher2001deformation,clements2016evaluation,virga2018use}. For instance, inter-examination variations in contact force can introduce substantial measurement errors during nodule size assessment, potentially undermining diagnostic reliability. Therefore, deformable registration is critical for enhancing measurement accuracy and improving diagnostic outcomes. On the other hand, differences in tissue deformability offer valuable biomechanical information, enabling clinicians to distinguish morphologically similar structures and diagnose certain diseases. For example,  veins exhibit greater deformability than arteries under contact force, serving as a practical discrimination criterion \cite{geng2024force}. Accurate registration clearly reveals such patterns with deformation fields, thereby improving diagnostic precision. Moreover, pathological changes, such as breast cancer \cite{ricci2014clinical,sadigh2012accuracy} or thyroid nodules \cite{cantisani2015strain,cantisani2014ultrasound}, often lead to altered tissue stiffness. Capturing these changes via deformation field estimation offers valuable insights for early diagnosis and treatment planning. 

However, several inherent challenges of ultrasound imaging complicate accurate deformable registration. The low contrast and indistinct tissue boundaries in grayscale ultrasound images provide limited semantic information for deep neural networks to learn meaningful representations. In addition, ultrasound images are often degraded by heavy speckle noise and acoustic artifacts, which interfere with feature extraction and spatial alignment. 
Existing learning-based registration methods predominantly rely on image intensity similarity metrics without incorporating physical priors \cite{balakrishnan2019voxelmorph,chen2022transmorph,kim2022diffusemorph,kim2021cyclemorph}.  When dealing with large deformation caused by certain anatomical structures like veins, the registration performance will be unsatisfactory~\cite{zhao2024correspondence}. The intensity-based methods tend to confuse the correct anatomical structures with other structures with similar gray values. Although the warped image resembles the target image, the anatomical structure is completely destroyed. As a result, the computed deformation field lacks physical interpretability and anatomical consistency.
 
\begin{figure}[!t]
\centerline{\includegraphics[width=\columnwidth]{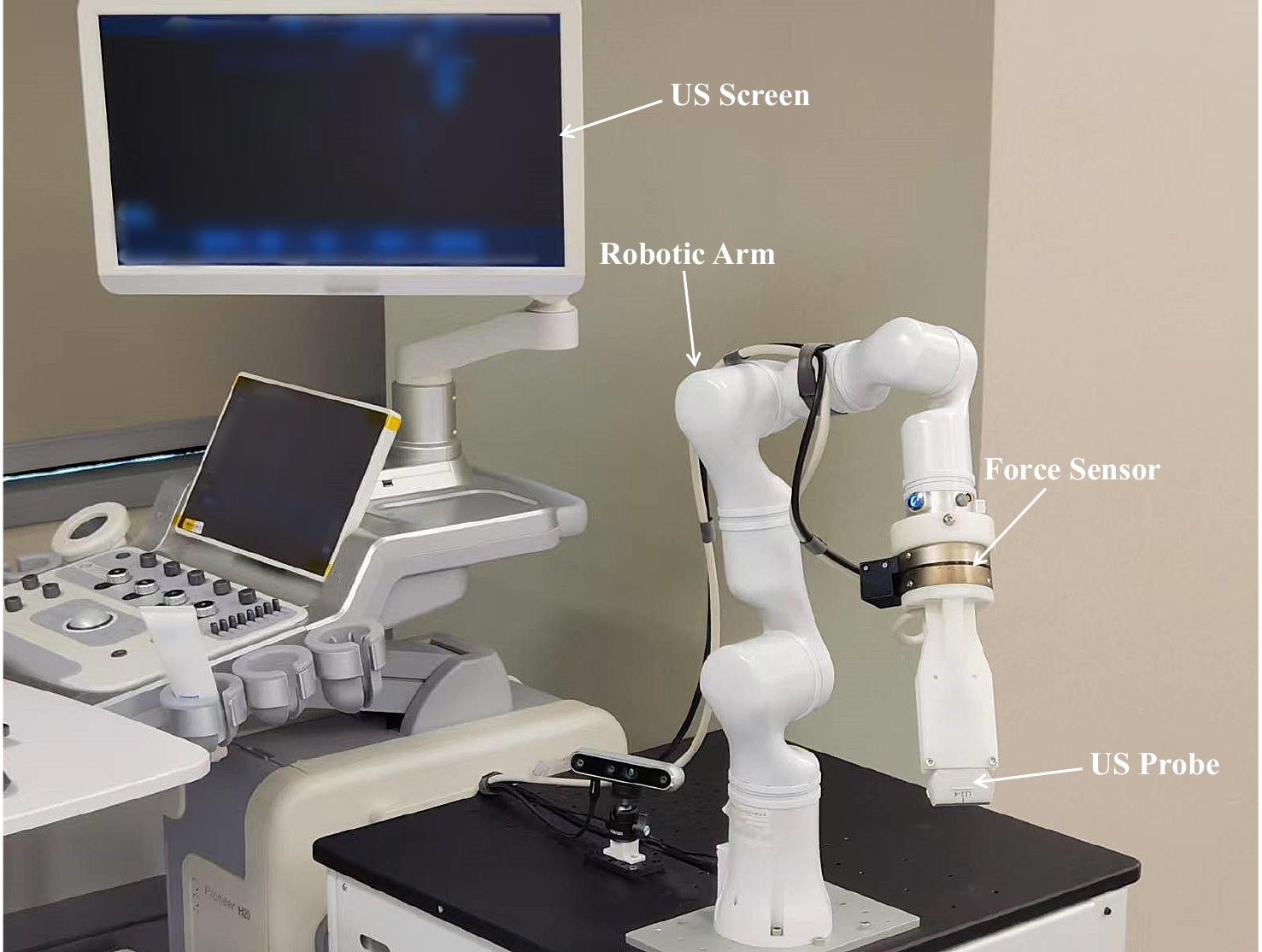}}
\caption{An overview of the robotic ultrasound system used in our study. A robotic arm precisely controls the pose and position of the ultrasound (US) probe. A force sensor is integrated into the probe to capture real-time contact force during scanning. Ultrasound images are monitored in real time via the US screen.}
\label{fig2}
\end{figure}

The emergence of Robotic Ultrasound Systems (RUSS) presents an opportunity to address these limitations. As illustrated in Fig. \ref{fig2}, RUSS integrated high-precision force sensors enable quantitative measurements of probe-tissue contact force during image acquisition. Moreover, both the direction and the magnitude of tissue deformation are directly related to the contact force.  This opens up the possibility of integrating force information to inform and constrain deformable registration.

Inspired by this, we propose the first physics-aware registration framework that integrates contact force measurements with image data, PADReg, to generate physically constrained deformation fields. To bridge the gap between contact force data and ultrasound image pairs, we introduce a pixel-wise tissue stiffness map that characterizes the deformability properties of anatomical structures. Rather than constructing patient-specific or organ-specific physical models, we leverage a deep neural network to directly infer the stiffness map from paired force and image data. To effectively extract informative features from the low-dimensional contact force input, we design a force encoder that projects the force signals into high-dimensional embeddings using sinusoidal positional encoding (sine and cosine functions). These force embeddings are subsequently fused with image embeddings to form unified multi-modal representations. To ensure physical plausibility, we further propose a physics-aware deformation estimation module that explicitly models the relationship between the learned stiffness map and the resulting deformation field. Finally, a spatial transformation is applied to generate the warped image. The loss is computed between the warped image and the target image, enabling end-to-end optimization of the entire framework via backpropagation.

The main contributions of the paper are: 
(1) We introduce the first approach to leverage contact force data for estimating large deformation fields in ultrasound deformable registration.
(2) We design an efficient encoder for force modality to enable multi-modal fusion.
(3) We propose a physical-aware registration framework inspired by the physical principle, thereby providing physical plausibility in deformation fields.
(4) Extensive experiments demonstrate that our method outperforms existing state-of-the-art ultrasound registration approaches.

\section{Related Work}

\subsection{Robotic Ultrasound Systems}
Robotic ultrasound systems (RUSS) have emerged as a significant advancement in medical imaging, offering enhanced precision, reproducibility, and reduced operator dependency compared to traditional manual ultrasound techniques \cite{jiang2023robotic}. These systems typically consist of several key components: a robotic manipulator, an ultrasound probe, a force sensor, and supplementary sensing modules (Fig. \ref{fig2}). To reduce the risk of transmitting pandemics, Ye \textit{et al}. \cite{ye2021feasibility} developed a 5G-Based robot-assisted remote US system for Covid-19 diagnosis, demonstrating its clinical feasibility. Suligoj et al. \cite{suligoj2021robust} implemented a high-level autonomous robotic framework for medical US imaging, leveraging 3D camera-derived patient-specific 2D skin surface to guide probe navigation. Jiang \textit{et al}. \cite{jiang2021autonomous} advanced the field by training a U-Net neural network for real-time vascular structure segmentation in ultrasound images. Their methodology combined 3D point cloud processing and nonlinear optimization techniques to enable fully automated probe positioning and tubular structure screening, with additional capability for accurate estimation of target vessel radius.
\subsection{Deformable Image Registration}
Recent progress in the ﬁeld of deep learning has signiﬁcantly advanced the performance of deformable medical image registration. Learning-based registration methods can be classified into three categories: fully-supervised methods, unsupervised methods, and weakly-supervised methods. 
Fully-supervised methods use the supervision of real deformation ﬁelds generated from traditional registration or synthetic deformation ﬁelds obtained from statistical models or random transformations. Fu \textit{et al}. \cite{fu2021biomechanically} utilized population-based finite element models from point clouds to degenerate the supervision deformation and designed an MR-TRUS registration network. Uzunova \textit{et al}. \cite{uzunova2017training} employed FlowNet \cite{dosovitskiy2015flownet} architecture for  2D MRI registration. Their ground truth is generated by statistical appearance models. 
Unsupervised methods solve the limitations of supervised methods in generating the required ground truth and generalizing results in different domains~\cite{zhao2025occlusion}. As a milestone, Balakrishnan \textit{et al} \cite{balakrishnan2019voxelmorph}. proposed Voxelorph, an end-to-end unsupervised U-Net framework. The loss function minimizes the difference between the warped image and the target image based on image intensity metrics.
TransMorph \cite{chen2022transmorph} is a hybrid Transformer-ConvNet framework utilizing Swin Transformer \cite{liu2021swin} as the encoder and trained end-to-end. Weakly-supervised registration methods build upon unsupervised registration methods by incorporating labels as auxiliary supervision. Hu \textit{et al}. \cite{hu2018weakly} utilized image pairs with multiclass segmentation labels to train an end-to-end CNN for 3D MRI and US image registration. Xu \textit{et al}. \cite{xu2019deepatlas} proposed DeepAtlas, which jointly learned networks for image registration and segmentation through loss of anatomy similarity.

\subsection{Ultrasound Elastography}
Ultrasound elastography (USE) is a non-invasive imaging technique that meatures changes in tissue elasticity to aid in disease diagnosis and characterization. 
It operates on the principle that pathological tissues, such as tumors \cite{wang2013viscoelastic} or fibrotic lesions \cite{cantisani2015strain}, exhibit altered mechanical properties compared to healthy tissues. USE is widely used in liver fibrosis assessment \cite{herrmann2018assessment}, breast lesion characterization \cite{ricci2014clinical,sadigh2012accuracy}, and thyroid nodule evaluation \cite{cantisani2015strain,cantisani2014ultrasound}. Meanwhile, USE also shows potential in vascular\cite{li2022arterial}, renal \cite{onur2015utility,menzilcioglu2015strain}, prostate \cite{junker2014real}, and lymph node imaging \cite{xu2011eus}. 

\begin{figure*}[htbp]
\centerline{\includegraphics[width=\textwidth]{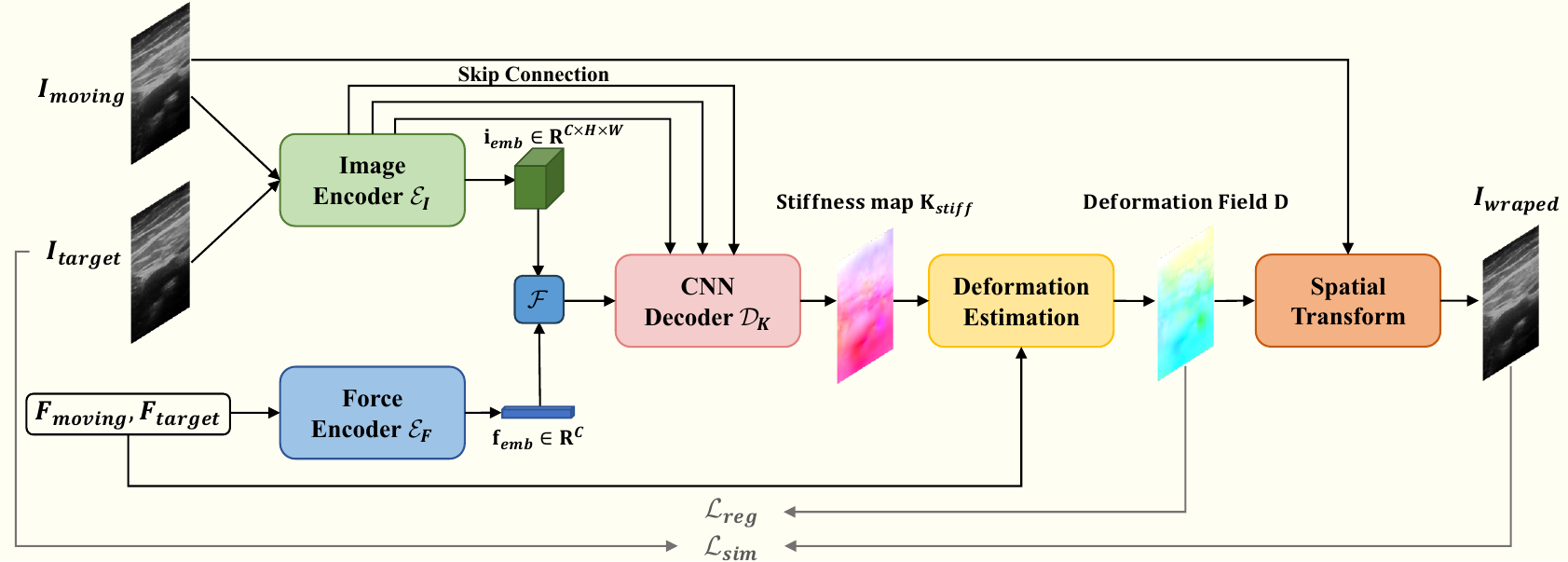}}
\caption{An overview of the proposed physics-informed deformation field estimation framework. The network takes a pair of ultrasound images $(\mathbf{I}_{moving}, \mathbf{I}_{target})$ and its corresponding contact force pair $(F_{moving},F_{target})$ as input. Visual features are extracted using an image encoder $\mathcal{E}_I$, while contact force signals are embedded using a force encoder $\mathcal{E}_F$. The image and force embeddings are fused via a feature fusion module $\mathcal{F}$ and passed to a CNN decoder $ \mathcal{D}_{K}$ with skip connections to generate a stiffness map $\mathbf{K}_{stiff}$. This map guides the deformation estimation module to produce the final deformation field $\mathbf{D}$, which is then used by the spatial transform module to generate the registered image $\mathbf{I}_{wrapped}$. The model is trained using a similarity loss $\mathcal{L}_{sim}$ and a regularization loss $\mathcal{L}_{reg}$.}
\label{fig3}
\end{figure*}

Strain elastography (SE) is a mainly used USE method, which measures tissue displacement under external stress. The performance of SE heavily depends on the accuracy of displacement estimation. Model methods generate the deformation based on radiofrequency (RF), Doppler processing, or a combination of the two methods \cite{sigrist2017ultrasound}. Ashikuzzaman \textit{et al.} \cite{ashikuzzaman2019global} developed a a novel quasi-static technique named GUEST, which optimizes a cost function based on three RF frames, incorporating spatial and temporal regularizations. Learning-based methods employ neural networks to compute elasticity images. Tehrani \textit{et al.} \cite{tehrani2022bi} proposed a semi-supervised method in which an optical flow network pretrained on computer vision images are fine-tuned using ultrasound data. 

Notably, several studies have leveraged concepts related to ultrasound elastography for various ultrasound analysis tasks. For instance, the DefCor-Net\cite{jiang2023defcor} framework introduced the concept of stiffness from elastography to construct patient-specific stiffness maps, which were subsequently used for deformation recovery of ultrasound images. Jiang et al. \cite{jiang2021deformation} utilized the contact force and robot pose data to build patient-specific models for the correction of 3D ultrasound volumes, which required palpation to be performed for each individual patient. Additionally, the UltRAP-Net\cite{li2025ultrap} proposed a network that predicts tissue properties from ultrasound image sequences and demonstrated promising performance.

\section{Methodology}

\subsection{Overview}

Our work architecture is based on the regular framework of learning-based registration, in which the network takes a pair of two deformed ultrasound images (i.e., a moving image $\mathbf{I}_{moving}$ and a target image $\mathbf{I}_{target}$) as input and generates a deformation field as final output. However, instead of directly generating the deformation field by the decoder like existing methods \cite{balakrishnan2019voxelmorph,chen2022transmorph}, our proposed method first generates a pixel-wise stiffness map showing the different deformability of different anatomical structures (As shown in Fig. \ref{fig3}). 

Then the stiffness map is combined with the force data through the physics-aware field estimation module to generate final deformation fields. The contact force,  recorded by a probe-mounted haptic sensor, is a scalar value along the axis normal to the skin surface and is temporally aligned with each ultrasound frame. Consequently, an image pair ($\mathbf{I}_{moving}$, $\mathbf{I}_{target}$) corresponds to a pair of force data ($F_{moving}$,  $F_{target}$). This force pair is also encoded through the force encoder and fused with image features to leverage multi-modal information, enhancing the registration performance.

\subsection{Prior-Knowledge-Free Stiffness Map Generation}
Most previous studies about ultrasound elastography estimate tissue stiffness based on finite element (FE) models \cite{fu2021biomechanically,samei2018real} or patient-specific prior information\cite{jiang2023defcor, jiang2021deformation}. In contrast, our method leverages the strong representation capacity of deep networks and intrinsic multi-modal information to predict pixel-wise stiffness maps directly from image pairs and synchronized force measurements. By learning stiffness maps in a fully data-driven manner, our approach naturally adapts to anatomical and biomechanical variations, eliminating the need for any prior knowledge or manual modeling assumptions required in previous methods. 

The input of the framework includes an image pair $(\mathbf{I}_{moving}, \mathbf{I}_{target})$ and a related force pair $(F_{moving},F_{target})$. The generated deformation field $\mathbf{D}\in \mathbb{R}^{2\times H\times W}$ is used to transform $\mathbf{I}_{moving}$ to $\mathbf{I}_{target}$ using a warping function, where $H$ and $W$ respectively stand for the height and width of input ultrasound images.
Following \cite{chen2022transmorph}, we employ the hybrid architecture including a Transformer-based image encoder $\mathcal{E}_{I}$ and a CNN-based decoder $\mathcal{D}_{K}$ to predict the stiffness map $\mathbf{K}_{stiff}$, as illustrated in Fig. \ref{fig3}. 
Transformer-based encoder has the ability to model long-range dependencies through global registration, and CNN-based decoder demonstrate strong capability in recovering local features. 

However, image pairs alone are insufficient for accurate stiffness estimation. Therefore, we introduce a novel force-visual fusion strategy, integrating force features with visual features before the decoder. In order to transfer the force data, which is only a scalar for each ultrasound image, to high-dimensional features, a force encoder $\mathcal{E}_{F}$ is used to align the dimensions between two modalities. The architecture of $\mathcal{E}_F$ is further demonstrated in Sec. \ref{sec:fe}. 
In conclusion, the process of generating the stiffness map $\mathbf{K}_{stiff}$ can be described formally in Eqs. \eqref{eq1}-\eqref{eq4}.
\begin{gather}
    \mathbf{i}_{emb} = \mathcal{E}_{I}[(\mathbf{I}_{moving},\mathbf{I}_{target})] \label{eq1}\\
    \mathbf{f}_{emb} = \mathcal{E}_{F}[(F_{moving},F_{target})]\\
    \mathbf{i}_{fused} = \mathcal{F}(\mathbf{i}_{emb},\mathbf{f}_{emb}) \\
    \mathbf{K}_{stiff} = \mathcal{D}_{K}[\mathbf{i}_{fused}] \label{eq4}
\end{gather}
where $\mathcal{F}$ represents the fusion function of the visual feature $\mathbf{i}_{emb}$ and the force feature $\mathbf{f}_{emb}$. $\mathbf{i}_{fused}$ represents the fused multi-modal features. For computational efficiency,  $\mathbf{f}_{emb}$ is expanded and subsequently multiplied with $\mathbf{i}_{emb}$ in a point-wise manner, which is proved to be effective through experiments.

\subsection{Contact Force Embedding} \label{sec:fe}
\begin{figure}[!t]
\centerline{\includegraphics[width=\columnwidth]{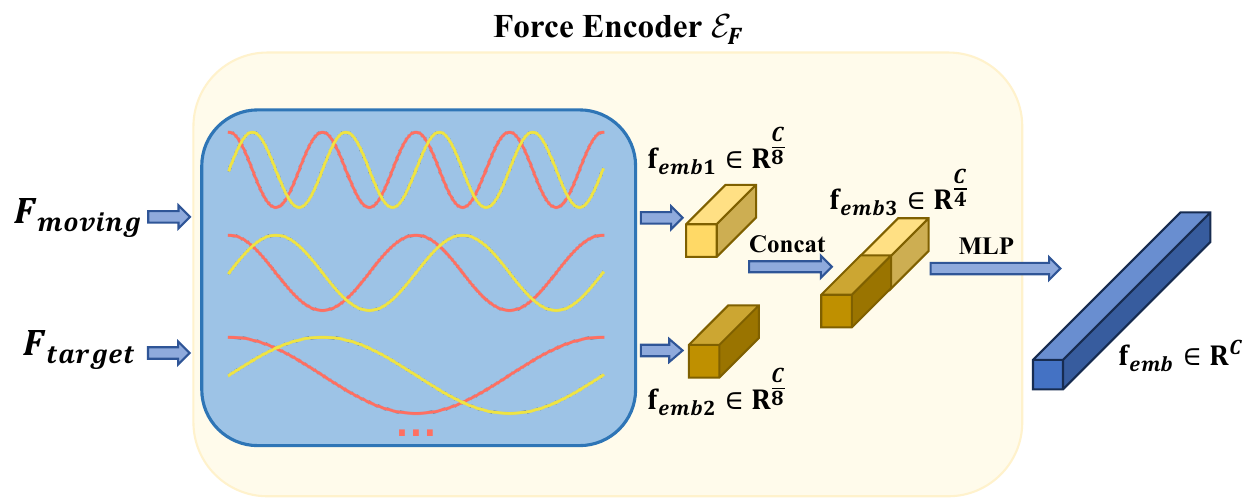}}
\caption{The illustration of the force encoder $\mathcal{E}_F$. The orange and yellow curves within the blue box represent sine and cosine functions at different frequencies. $C$ denotes the dimension of visual features.}
\label{fig4-1}
\end{figure}

During ultrasound data acquisition, a force sensor integrated into the probe records the real-time contact force at a sampling frequency higher than the ultrasound video frame rate. To reduce computational redundancy, each ultrasound frame is temporally aligned with a corresponding force measurement based on their respective timestamps. 

Meanwhile, the control method proposed in \cite{cao2023ultra} is adopted to maintain the ultrasound probe perpendicular to the skin surface. This vertical alignment maximizes acoustic wave transmission and  thus enhances imaging quality. Given the probe's consistent perpendicular orientation, only the force component along the probe's Z-axis (i.e., normal to the skin surface) is considered in subsequent analyses. Furthermore, the direction of the contact force  remains constant, which is always oriented toward the patient's skin.

Consequently, the force data $F$ associated with the image frame $\mathbf{I}$ is reduced to a non-negative scalar quantity, i.e., $F \in \mathbb{R}_{\geq0}$. 
Inspired by the positional encoding strategy in Transformers \cite{vaswani2017attention}, the originally scalar contact force is projected into a high-dimensional embedding space to effectively capture both the magnitude and direction of tissue compression, facilitating seamless fusion with visual features. 

The encoding process for the contact force data is illustrated in Fig. \ref{fig4-1}. Specifically, each scalar force value $F$ is encoded using sine and cosine functions with varying frequencies to generate its embedding representation. The embedding comprises two components $\mathbf{f}_{embsin}$ and $\mathbf{f}_{embcos}$. Given a target embedding dimension $d_{model}$, the $j$-th dimension (where $j\in [0,\frac{d_{model}}{2})$) is computed as follows:
\begin{align}
\mathbf{f}_{embsin}(F,j)=\sin{\left(\frac{F}{1000^{2j/d_{model}} }\right) } \label{eqC1}\\
\mathbf{f}_{embcos}(F,j)=\cos{\left(\frac{F}{1000^{2j/d_{model}} }\right)} \label{eqC2}
\end{align}

Since each input image pair corresponds to two force values $F_{moving}$ and $F_{target}$, we independently compute their corresponding embeddings:
\begin{align}
\mathbf{f}_{emb1} &= \text{concat}(\mathbf{f}_{embsin}(F_{moving}),\mathbf{f}_{embcos}(F_{moving})) \label{eqC3}\\
\mathbf{f}_{emb2} &= \text{concat}(\mathbf{f}_{embsin}(F_{target}),\mathbf{f}_{embcos}(F_{target})) \label{eqC4}
\end{align}
 where $\mathbf{f}_{emb1}, \mathbf{f}_{emb2} \in  \mathbb{R}^{d_{model}}$. These two vectors are then concatenated together to form a joint force embedding  with higher dimension:
\begin{equation}
    \mathbf{f}_{emb3} = \text{concat}(\mathbf{f}_{emb1},\mathbf{f}_{emb2}).
\end{equation}
where $\mathbf{f}_{emb3} \in \mathbb{R}^{2d_{model}}$.  

To align the dimension with the visual embedding $\mathbf{i}_{emb} \in \mathbb{R}^{C}$, two additional linear layers are applied to project  $\mathbf{f}_{emb3}$ into the final force embedding space $\mathbf{f}_{emb} \in \mathbb{R}^{C}$. In practice, we set $d_{model} = \frac{C}{8}$.

\subsection{Physics-aware Deformation Field Estimation}
According to the Hooke’s Law, object strain is proportional to the applied stress, expressed as:
\begin{equation}
     \epsilon = \sigma/E
     \label{hookes}
\end{equation}
where $\sigma$ denotes stress, $\epsilon$ represents the resulting strain, and $E$ stands for the Young’s modulus of the object.

In our quasi-static ultrasound setting, assuming a fixed probe orientation and contact area, the contact force can be considered a surrogate for stress.
Therefore, we model the deformation field as directly proportional to the force difference $\Delta F$ and inversely related to the stiffness. The predicted stiffness map $\mathbf{K}_{stiff}$ is a learnable proxy for the inverse of local Young’s modulus, allowing the network to encode tissue mechanical properties in a pixel-wise manner. Regions with lower stiffness are expected to undergo greater deformation for a given force, while stiffer regions deform less. Nevertheless, due to the intrinsic imaging mechanism of ultrasound, the actual tissue deformations are significantly more complex. We simplify the problem in order to ensure the feasibility and computational efficiency of model training.

\begin{figure}[!t]
\centerline{\includegraphics[width=\columnwidth]{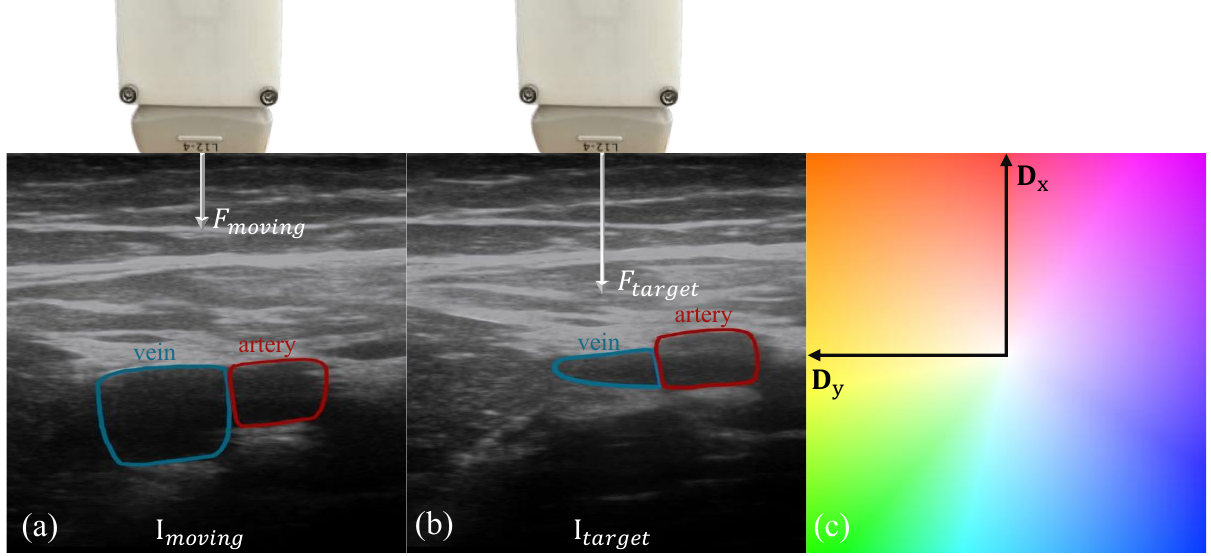}}
\caption{(a) The moving image $I_{moving}$ acquired with  $F_{moving}$ applied on the to the skin surface. (b) The target image   $I_{target}$ obtained under $F_{target}$, where $F_{target} >F_{moving}$. This force differential induces upward tissue deformation and compression. (c) Illustration of  flow color encoding and  the deformation coordinate system used in the article. Under the condition $F_{target} >F_{moving}$, $D_x \succ 0$ is expected for most pixel regions. }
\label{fig4}
\end{figure}

As illustrated in Fig. \ref{fig4}, $\mathbf{D}_x$ maintains a consistent directional relationship with $\Delta F$, where increased force (i.e., $\Delta F > 0$) causes upward displacement (i.e.,  all elements of  $\mathbf{D}_x$ are positive, $\mathbf{D}_x \succ 0$ ) and decreased force (i.e., $\Delta F < 0$) leads to downward motion (i.e., all elements of  $\mathbf{D}_x$ are negative, $\mathbf{D}_x \prec  0$). Meanwhile, although $\mathbf{D}_y$ has no direct directional relationship with $\Delta F$, its limited magnitude enables accurate prediction by the network.

The resulting deformation field is computed as follows:
\begin{align}
    \mathbf{D}_x = \mathbf{K}_{stiffx}\times \Delta F \label{eqD1}\\
    \mathbf{D}_y = \mathbf{K}_{stiffy}\times \Delta F \label{eqD2}
\end{align}
where $\mathbf{K}_{stiffx} \in \mathbb{R}^{H\times W}$ and $\mathbf{K}_{stiffy} \in \mathbb{R}^{H\times W}$ represent the stiffness map components along the X-axis and Y-axis, respectively. $\mathbf{D}_x \in \mathbb{R}^{H\times W}$ and $\mathbf{D}_y \in \mathbb{R}^{H\times W}$ denote the deformation field components along respective axes. The final deformation field is formed by concatenating these components: $\mathbf{D}=\left [\mathbf{D}_x,\mathbf{D}_y  \right ]$. This physics-inspired relationship enables the resulting deformation field to reflect real tissue mechanics more accurately.

Using raw force differences directly $F_{target}-F_{moving}$ can result in poor generalization, 
as absolute force magnitudes exhibit large variability between patients and regions.
We therefore introduce a normalized force formulation to capture relative changes in contact force more robustly:
\begin{equation}
    \Delta F = \text{sign}(F_{target}-F_{moving})\sqrt{\left | \frac{F_{target}-F_{moving}}{F_{target}+F_{moving}}\right |}.
\end{equation}
This formulation normalizes the force difference by the total applied force, ensuring that deformation responses are scale-invariant across different patients or probe settings. Empirical results (Table \ref{table:ab2}) confirm its superiority over alternative designs.

The loss function of our self-supervised framework comprises a similarity loss $\mathcal{L}_{sim}$ and a regularization loss $\mathcal{L}_{reg}$.  The spatial transform module generates a warped image $\mathbf{I}_{warped}$ from $\mathbf{I}_{moving}$ using the deformation field $\mathbf{D}$. The similarity loss $\mathcal{L}_{sim}$ minimizes the intensity difference between $\mathbf{I}_{warped}$ and $\mathbf{I}_{target}$:
\begin{equation}
    \mathcal{L}_{sim} = MSE(\mathbf{I}_{warped},\mathbf{I}_{target})
\end{equation}

The regularization loss $\mathcal{L}_{reg}$ penalizes abrupt changes in the deformation field,  imposes smoothness in the deformation field \cite{balakrishnan2019voxelmorph}:
\begin{equation}
    \mathcal{L}_{reg} = \sum_{\mathbf{p}\in \Omega}\left \| \nabla \mathbf{D}(\mathbf{p} ) \right \| 
\end{equation}
where $\Delta\mathbf{D}\left(\mathbf{p} \right)$ represents the spatial gradients of the deformation field at point $\mathbf{p}$. The gradient is approximated via forward differences:
\begin{equation}
    \Delta\mathbf{D}(\mathbf{p} ) \approx  \mathbf{D}\left(p_{\{x,y\}}+1\right)-\mathbf{D}\left(p_{\{x,y\}}\right)
\end{equation}

\section{Experiments And Results}
\begin{table*}[htbp]
\caption{Comparison Results of Different Methods on the Mus-V Dataset.}
\label{tabel:1}
\setlength{\extrarowheight}{3pt} 
\begin{tabularx}{\linewidth}{c
>{\centering\arraybackslash}X
>{\centering\arraybackslash}X
>{\centering\arraybackslash}X
>{\centering\arraybackslash}X
>{\centering\arraybackslash}X
>{\centering\arraybackslash}X} 
\hline
        Model & DSC$\uparrow$ & HD95$\downarrow$  & SSIM$\uparrow$   & MSE$\downarrow$ & MI$\downarrow$ & DR(\%)$\downarrow$ \\ \hline
        Voxelmorph\cite{balakrishnan2019voxelmorph} & 0.5074 & 21.08 & 0.6335  & 0.0044 & -0.5510 & 12.29 \\
        Transmorph\cite{chen2022transmorph} & 0.6332 & 16.40 & 0.6054 & 0.0054  & -0.5278 & 11.16 \\ 
        Transmorph-diff\cite{chen2022transmorph} & 0.2308 & 38.69 & 0.4668  & 0.0150 & -0.2410 & 50.00 \\ 
        Transmorph-Bayes\cite{chen2022transmorph} & 0.5791 & 19.05 & 0.6735  & 0.0037 & -0.5923 & 8.11 \\ 
        Transmorph-bspl\cite{chen2022transmorph} & 0.6216 & 26.52 & 0.6731  & 0.0037 & -0.5925 & 5.90 \\ 
        Diffusemorph\cite{kim2022diffusemorph} & 0.6861 & 23.81 & 0.6698  & 0.0057 & -0.5412 & 2.58 \\
        Cyclemorph\cite{kim2021cyclemorph} & 0.3436 & 25.13 & 0.6773  & 0.0071 & -0.4688 & 27.86 \\ 
        SYN\cite{avants2014insight} & 0.6476 & Inf & 0.6491 & 0.0104 &-0.6079 & 50.77 \\
        TVMSQC\cite{avants2014insight} & 0.5672 & 17.85 & 0.6134  & 0.0055 & -0.5186 & 11.50 \\ 
         \hline
        Ours & \textbf{0.7146} & \textbf{12.90} & \textbf{0.6856} & \textbf{0.0036} & \textbf{-0.6093} & \textbf{2.54} \\ \hline
\end{tabularx}
\end{table*}

\begin{figure*}[htbp]
\centerline{\includegraphics[width=\textwidth]{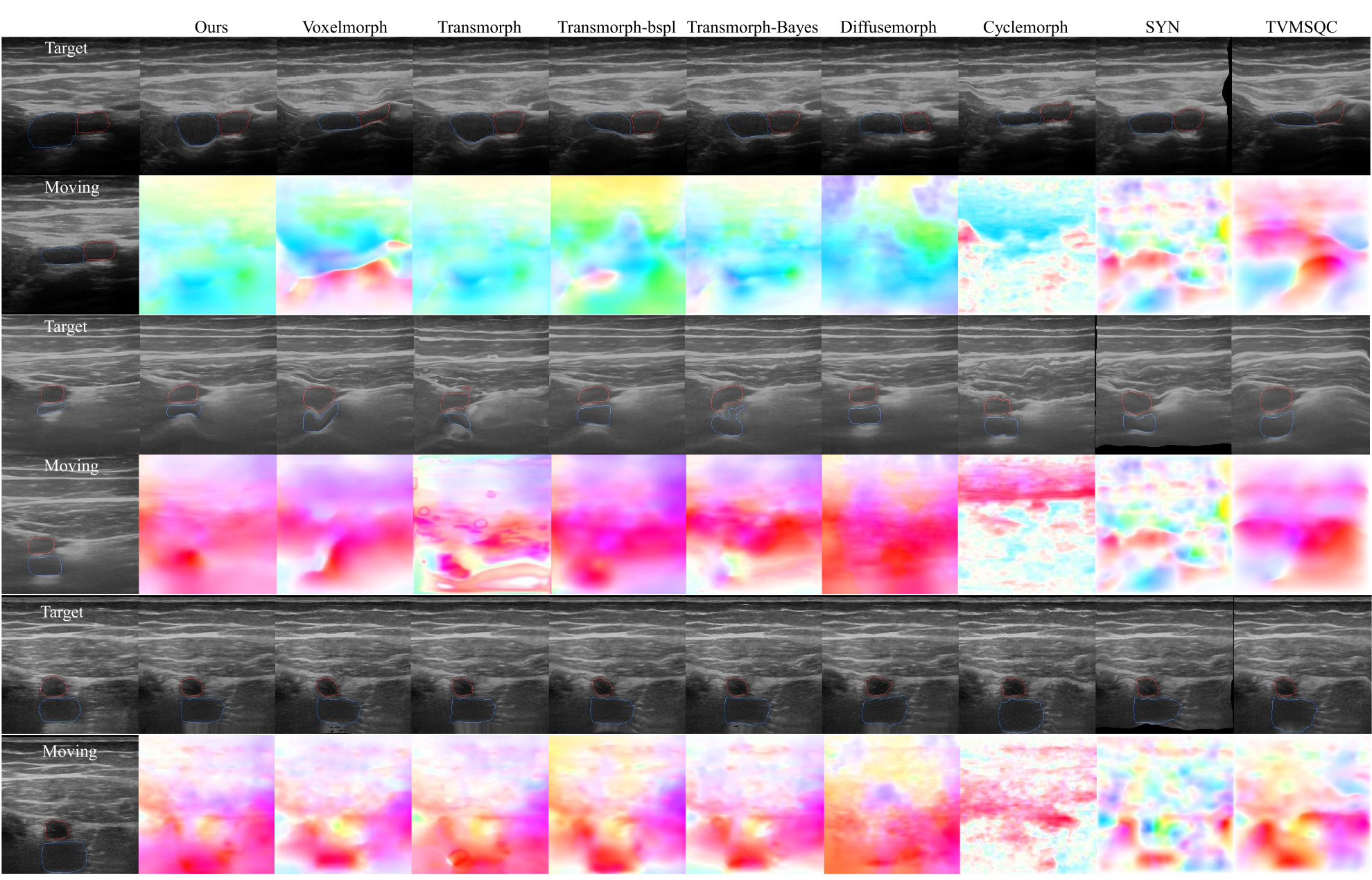}}
\caption{Visualization results. Arteries and veins are labeled in red and blue, respectively. Color encoding pattern is represented in Fig. \ref{fig4}(c). Visualization results. Arteries and veins are labeled in red and blue, respectively. The color encoding scheme is illustrated in Fig. \ref{fig4}(c). As shown in the deformation fields, our method effectively constrains the deformation toward the correct direction, thereby substantially enhancing the physical interpretability of the deformation fields.}
\label{fig5}
\end{figure*}

\subsection{Dataset and Preprocessing}
The Mus-V\cite{geng2024force} dataset was employed to validate the performance of the proposed network. This multi-modal dataset comprises both real ultrasound video sequences and corresponding contact force measurements. The ultrasound dataset is collected from 11 healthy volunteers with diverse characteristics, including variations in gender and body mass index (BMI). All sequences were acquired using a standardized Angell Pioneer H20 ultrasound scanner. 
We selected 1,926 high-quality femoral vascular images exhibiting complete anatomical structures without significant acoustic shadowing artifacts, comprising 1,324 training images and 602 validation images. Training and validation pairs were constructed by sampling frames from identical video sequences. To explicitly evaluate large-deformation handling capability of the network, we selected image pairs with force differentials exceeding 2N, which is an empirically determined threshold where force differences beyond this value induce substantial tissue deformation. This process yielded 16,094 training pairs and 10,425 validation pairs.

\subsection{Evaluation Metrics}
\begin{figure*}[!t]
\centerline{\includegraphics[width=\textwidth]{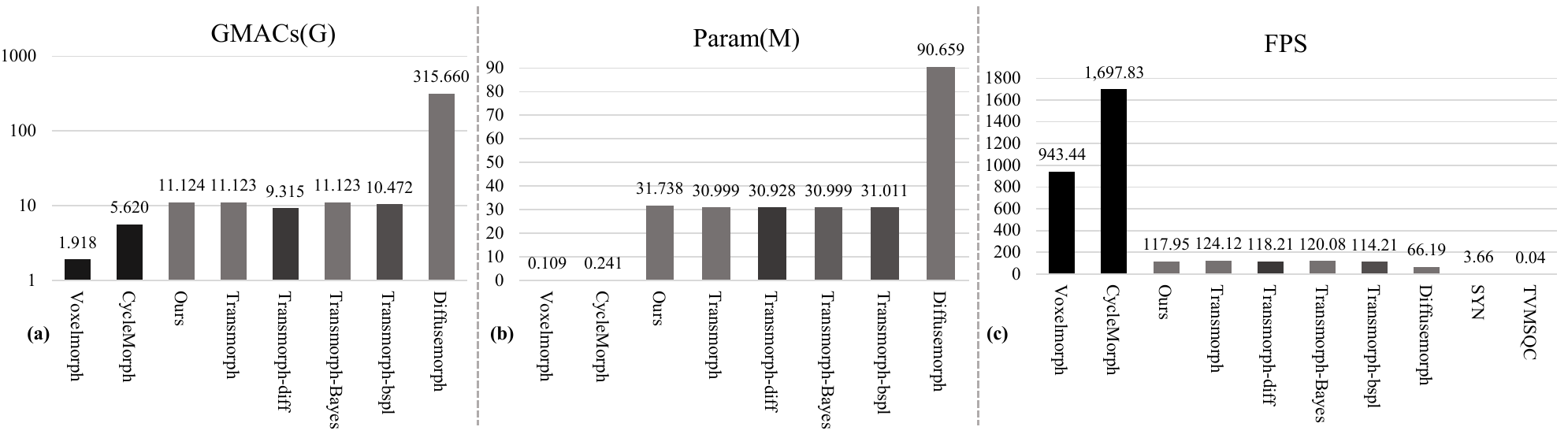}}
\caption{(a) The computation complexity comparisons represented in Giga multiply–accumulate operations (GMACs) (b) The comparisons of model parameters number. (c) The comparisons of prediction speed represented in frame per second (FPS). All the values are computed under the input size of 256x256.}
\label{fig6}
\end{figure*}

The performance of registration models was evaluated through comprehensive metrics, including structural metrics, intensity-based metrics and physics-based metrics. For structural metrics, the Mus-V dataset provides the segmentation masks for arteries and veins, so we employed Dice Similarity Coefficient (DSC) and 95th percentile Hausdorff Distance (HD95) to evaluate the consistency of anatomical structures. DSC measures the spatial overlap between the masks in warped and target images. HD95 measures the 95th percentile of the distances between the boundaries of the warped and target masks, indicating the extent of boundary mismatches while reducing the sensitivity to outliers. For intensity-base metrics, mono-modal registration is typical measured by Structure Similarity Index Measure (SSIM). Moreover, Mutual Information (MI) and MSE were also used to evaluate the similarity of $\mathbf{I}_{warped}$ and $\mathbf{I}_{target}$. 

Futhermore, to evaluate the physical plausibility of the deformation fields, we introduce a Discrepancy Rate (DR) metric based on the directional alignment between the X-component of the deformation field ($\mathbf{D}_x$) and the contact force difference $\Delta F$. The DR is computed as Eqs. \eqref{eq:DR1}-\eqref{eq:DR2}.
\begin{align}
    DR &= \frac{1}{HW}\sum^{H}_{i=1}\sum^{W}_{j=1}\mathbb{I}\left[\text{sign}\left(\mathbf{D}_x(i,j)\right) \ne \text{sign}(\Delta F)\right]
    \label{eq:DR1}
    \\
    &= \frac{1}{HW}\sum^{H}_{i=1}\sum^{W}_{j=1}\mathbb{I}\left[\mathbf{D}_x(i,j)\cdot \Delta F < 0 \right]
    \label{eq:DR2}
\end{align}
where $\mathbb{I}$ denotes the indicator function,  and $H$ and $W$ represent the height and width of the deformation field, respectively.

\subsection{Implementation Detail}
The proposed network was implemented using PyTorch and trained on a workstation with an NVIDIA A100 Tensor Core GPU. All models were trained for 100 epochs using the Adam optimizer with a learning rate of $1 \times 10^{-3}$ and a batch size of 32. We employe a step scheduler with a step size of 30 and a multiplicative factor of 0.7 was used. The loss weights for $\mathcal{L}_{sim}$ and $\mathcal{L}_{reg}$ were set to 1 and 0.03, respectively. Input images were resized to $256 \times 256$ pixels during both training and validation.

\subsection{Comparison with State-of-the-Art Methods}
We compared our approach against prominent deformable registration methods, including Voxelmorph \cite{balakrishnan2019voxelmorph} and Transmorph \cite{chen2022transmorph} (with its variants: Transmorph-diff, Transmorph-Bayes and Transmorph-bspl). 
Recent techniques such as Diffusemorph \cite{kim2022diffusemorph}, CycleMorph \cite{kim2021cyclemorph} were also considered in the comparison. 
Moreover, we took non-deep learning methods into account, including Symmetric Normalization (SyN) \cite{avants2014insight} and TVMSQC \cite{avants2014insight} which is a time-varying diffeomorphism method with mean square metric for very large deformation. For these two methods, we use the implementation in the publicly available Advanced Normalization Tools (ANTs) software package \cite{avants2011reproducible}.

Quantitative results of these baselines and our method are presented in Table \ref{tabel:1}. 
The comparison results denote that our method achieves both the highest structure-based metrics and the best intensity-based metrics. 
Our method obtains a HD95 of 12.90, which is 3.50 (21.34\%) lower than the second-best method.
Our method also obtains the highest DSC of 0.7143 and the lowest Discrepancy Rate (DR)  of 2.54\%. DR is a significant metric that indicates the physical reasonability of generated deformation fields. 
As shown in Table \ref{tabel:1}, although Transmorph and TVMSQC achieve notable HD95, their DR of more than 11\% indicates that although output images exhibit grayscale similarity to target images, the underlying anatomical structures are significantly compromised. 
This can be further observed in the visualization results.

The visualization results are shown in Fig. \ref{fig5}. Due to its unsatisfactory performance, the visualization results of Transmorph-diff is neglected.  The flow color encoding pattern is shown in Fig.\ref{fig4}(c). As observed, our method generates the smoothest deformation fields and  predicts the correct direction for large deformation.

To evaluate the training cost and inference speed of the network, we assessed the computational complexity, parameter size, and inference FPS of our proposed method and all baselines (Fig.\ref{fig6}). Giga multiply–accumulate operations (GMACs) represent the number of multiply–accumulate operations performed per second by the model. A larger GMACs value indicates higher computational complexity. Frames Per Second (FPS) denotes the number of samples that the model can process per second during inference, with the sample size of 256×256. Since SyN and TVMQC are not deep learning-based methods, only FPS is reported for these two approaches. The results demonstrate that our method achieves a moderate computational complexity while maintaining sufficient real-time performance.

\subsection{Ablation Studies}
\subsubsection{Multi-Modal Fusion Effectiveness}
The proposed network encodes the force data and fuses the force features with the visual features. To validate the effectiveness of multi-modal fusion, we implemented an ablation model excluding the force encoder $\mathcal{E}_F$ and fusion function $\mathcal{F}$. Additionally, we explored several alternative designs of $\mathcal{F}$ to find the optimal structure. These alternative designs include the point-wise addition and the cross-attention mechanism where $\mathbf{I}_{emb}$ is flattened as keys/values and $\mathbf{f}_{emb}$ serves as queries. Results are summarized in Table \ref{table:ab1}. MI and MSE are neglected for simplicity. After removing the force embedding, the performance of the network decreases significantly, which is a strong proof of the effectiveness of the modal fusion strategy.  Moreover, simple point-wise addition and multiplication achieve better performance, rather than the more complex method using attention mechanism.

\begin{table}[htp]
\caption{Ablation Results for Multi-Modal Fusion.}
\label{table:ab1}
\setlength{\extrarowheight}{3pt} 
\begin{tabularx}{\columnwidth}{c
>{\centering\arraybackslash}X
>{\centering\arraybackslash}X
>{\centering\arraybackslash}X
>{\centering\arraybackslash}X}
\hline
        Model & DSC$\uparrow$ & HD95$\downarrow$  & SSIM$\uparrow$   & DR(\%)$\downarrow$ \\ \hline
        w/o $\mathbf{f}_{emb}$ & 0.6377 & 15.16 & 0.6279  & 2.65 \\
        Addition & 0.6944 & 13.64 & 0.6854  & 2.57 \\ 
        Attention & 0.6446 & 15.09 & 0.6522  & \textbf{2.54}  \\ \hline
        Ours  & \textbf{0.7146} & \textbf{12.90} & \textbf{0.6856}  & \textbf{2.54}\\ \hline
\end{tabularx}
\end{table}

\begin{figure}[htbp]
\centerline{\includegraphics[width=\columnwidth]{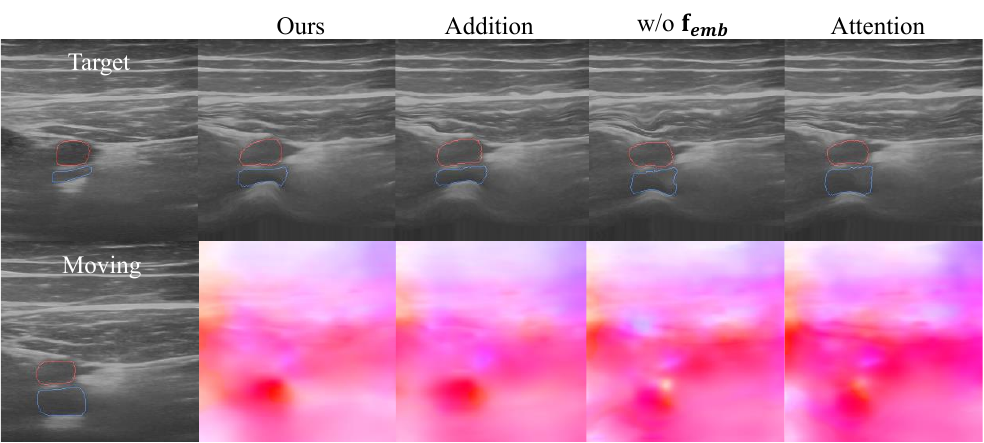}}
\caption{Visualization results of multi-modal fusion ablation study. Arteries and veins are labeled in red and blue, respectively.}
\label{fig7}
\end{figure}

Visualization results of different designs are shown in Fig.\ref{fig7}.
In the case of no feature fusion and attention mechanism, although the deformation fields maintain the correct direction, the magnitude of the deformation is not adequately captured.

\subsubsection{Contact Force Difference Formulation}
\begin{figure}[htbp]
\centerline{\includegraphics[width=\columnwidth]{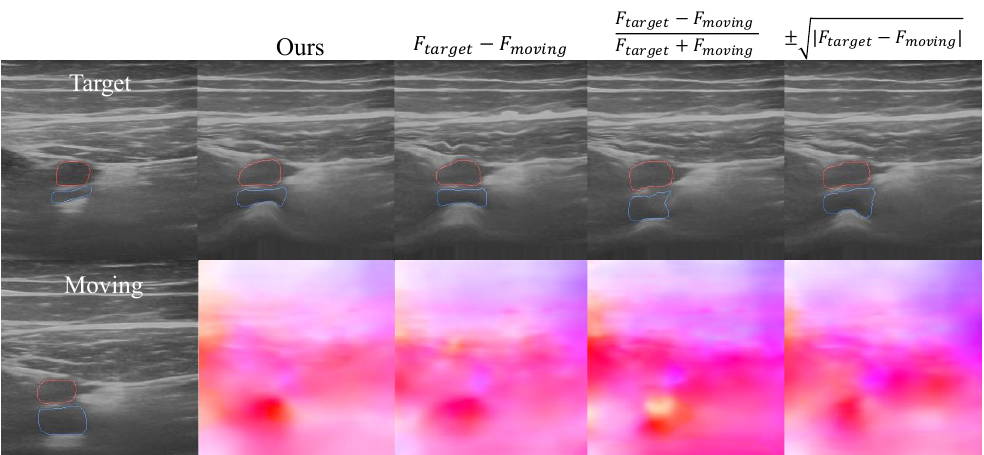}}
\caption{Visualization results of $\Delta F$ formulation ablation study. Arteries and veins are labeled in red and blue, respectively.}
\label{fig8}
\end{figure}

One of the most intuitive design of $\Delta F$ is $\Delta F=\sqrt{F_{target}}-\sqrt{F_{moving}}$. However, as shown in experimental results, this design leads to non-convergence of the network. In addition, we investigate two alternative designs for $\Delta F$, $ F_{target}-F_{moving}$ and $\frac{F_{target}-F_{moving}}{F_{target}+F_{moving}}$, and conduct corresponding experiments. Table \ref{table:ab2} presents the comparative results of several distinct $\Delta F$ calculation methods. The results also demonstrate that other calculation methods are inferior to our proposed method. Visualization results are shown in Fig.\ref{fig8}.

\begin{table}[htbp]
\caption{Ablation Results for $\Delta F$ Formulation}
\label{table:ab2}
\setlength{\extrarowheight}{4pt} 
\begin{tabularx}{\columnwidth}{c
>{\centering\arraybackslash}X
>{\centering\arraybackslash}X
>{\centering\arraybackslash}X
>{\centering\arraybackslash}X}
\hline
    $\Delta F$ & DSC$\uparrow$ & HD95$\downarrow$  & SSIM$\uparrow$   & DR(\%)$\downarrow$ \\ \hline
    $ F_{target}-F_{moving}$ & 0.6946 & 13.35 & 0.6779  & 2.93 \\
    $\frac{F_{target}-F_{moving}}{F_{target}+F_{moving}}$ & 0.6960 & 13.41 & 0.6809  & 2.79  \\ 
    $\pm \sqrt{\left |F_{target}-F_{moving}\right |}$ & 0.7016 & 13.35 & 0.6839 & 2.58 \\ 
    \hline
    Ours  & \textbf{0.7146} & \textbf{12.90} & \textbf{0.6856}  & \textbf{2.54}
    \\ \hline
\end{tabularx}
\end{table}

\subsubsection{Physics-Aware Deformation Field Estimation}
Instead of directly predicting the deformation field, our proposed method firstly predicts a stiffness map and then applies physical constraints to get final outcomes. To demonstrate the efficacy of this approach, we implemented an ablation network that directly predicts the deformation field without incorporating $\Delta F$, i.e., $\mathbf{D} = \mathbf{K}_{stiff}$.
In addition, we explored the design of two more sophisticated deformation field estimation modules. For the linear model, the deformation field components are given by:
\begin{align}
    \mathbf{D}_x = (\beta_x\mathbf{K}_{stiffx}+\alpha_x)\times \Delta F \\ \mathbf{D}_y = (\beta_y\mathbf{K}_{stiffy}+\alpha_y)\times \Delta F 
\end{align}
where $\alpha_x$, $\alpha_y$, $\beta_x$  and $\beta_y$ are scalar parameters.
For the quadratic model, the deformation field components are expressed as:
\begin{align}
    \mathbf{D}_x = (\gamma_x\mathbf{K}^{2}_{stiffx}+\beta_x\mathbf{K}_{stiffx}+\alpha_x)\times \Delta F\\
    \mathbf{D}_y = (\gamma_y\mathbf{K}^{2}_{stiffy}+\beta_y\mathbf{K}_{stiffy}+\alpha_y)\times \Delta F 
\end{align}
where $\gamma_x$ and $\gamma_y$ are additional scalar parameters. The experimental results are presented in Table \ref{table:ab3}. 
The results demonstrate that directly predicting deformation fields without incorporating $\Delta F$ leads to a prominent performance degradation. Additionally, 
our approach achieves better performance than linear and quadratic regression with lower computational complexity.

\begin{table}[htbp]
\caption{Ablation Results for Physics-Aware Deformation Field Estimation.}
\label{table:ab3}
\setlength{\extrarowheight}{3pt} 
\begin{tabularx}{\columnwidth}{c
>{\centering\arraybackslash}X
>{\centering\arraybackslash}X
>{\centering\arraybackslash}X
>{\centering\arraybackslash}X}
\hline
        Model & DSC$\uparrow$ & HD95$\downarrow$  & SSIM$\uparrow$   & DR(\%)$\downarrow$ \\ \hline
        w/o $\Delta F$ & 0.5034 & 22.00 & 0.6623  & 18.34 \\
        Linear model & 0.7086 & 13.11 & 0.6822  & 2.66 \\ 
        Quadratic model & 0.6941 & 13.43 & 0.6748  & 6.94  \\ \hline
       Ours  & \textbf{0.7146} & \textbf{12.90} & \textbf{0.6856}  & \textbf{2.54} \\ \hline
\end{tabularx}
\end{table}

\begin{figure}[htbp]
\centerline{\includegraphics[width=\columnwidth]{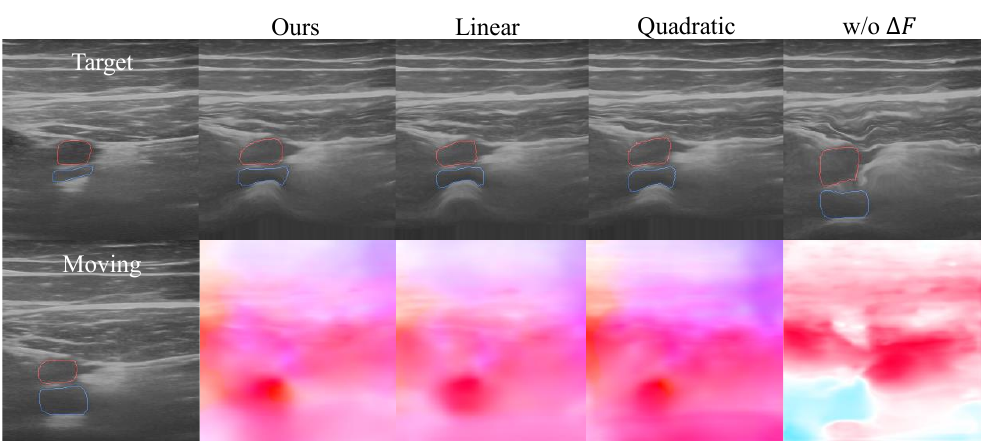}}
\caption{Visualization results of different deformation field prediction strategies. Arteries and veins are labeled in red and blue, respectively.}
\label{fig9}
\end{figure}

The visualization results (Fig.~\ref{fig9}) show that removing $\Delta F$ leads to even worse performance than the original TransMorph, indicating that the Physics-aware deformation field estimation module and  $\mathbf{f}_{emb}$ complement each other and work synergistically.

\section{Limitations and Future Work}
Despite the promising results, the proposed method has several limitations that warrant further investigation. First, the deformation mechanism has been simplified to facilitate efficient model training. Future work could focus on developing a more sophisticated network architecture that more accurately models the physical mechanism inherent in ultrasound imaging.

Second, due to the lack of multi-modal datasets, our experiments are conducted solely on the Mus-V dataset, which consists of healthy vascular images. As a result, the method's generalizability to pathological tissues or other anatomical regions has not been validated.  Broader datasets need to be constructed and validated to improve its generalizability in the future.

Another limitation is the reliance on robotic ultrasound systems equipped with force sensors, which may not be accessible in routine clinical practice. Future work could investigate force estimation directly from image features, reducing hardware dependency.

\section{Conclusion}
In  this work, we have presented a novel physics-aware multi-modal registration framework for handling large deformation in  ultrasound image sequences.  To the best of our knowledge, this work represents the first efforts to incorporate contact force data from robotic ultrasound systems as physical priors to improve the accuracy and interpretability of deformation field estimation. 

Our method achieved superior performance over existing methods in both anatomical consistency and physical plausibility. It also maintained a high inference speed, enabling real-time application. 

This study highlights that integrating physical constraints via contact force information significantly improves the performance of ultrasound registration. Moreover, the proposed concept of leveraging contact force as a form of physical prior has the potential to benefit a wider range of ultrasound image analysis tasks, including reconstruction, generation, segmentation, and recognition.

\bibliographystyle{IEEEtran}
\bibliography{ref}

\begin{thebibliography}{10}
\providecommand{\url}[1]{#1}
\csname url@samestyle\endcsname
\providecommand{\newblock}{\relax}
\providecommand{\bibinfo}[2]{#2}
\providecommand{\BIBentrySTDinterwordspacing}{\spaceskip=0pt\relax}
\providecommand{\BIBentryALTinterwordstretchfactor}{4}
\providecommand{\BIBentryALTinterwordspacing}{\spaceskip=\fontdimen2\font plus
\BIBentryALTinterwordstretchfactor\fontdimen3\font minus \fontdimen4\font\relax}
\providecommand{\BIBforeignlanguage}[2]{{%
\expandafter\ifx\csname l@#1\endcsname\relax
\typeout{** WARNING: IEEEtran.bst: No hyphenation pattern has been}%
\typeout{** loaded for the language `#1'. Using the pattern for}%
\typeout{** the default language instead.}%
\else
\language=\csname l@#1\endcsname
\fi
#2}}
\providecommand{\BIBdecl}{\relax}
\BIBdecl

\bibitem{sotiras2013deformable}
A.~Sotiras, C.~Davatzikos, and N.~Paragios, ``Deformable medical image registration: A survey,'' \emph{IEEE transactions on medical imaging}, vol.~32, no.~7, pp. 1153--1190, 2013.

\bibitem{ricci2014clinical}
P.~Ricci, E.~Maggini, E.~Mancuso, P.~Lodise, V.~Cantisani, and C.~Catalano, ``Clinical application of breast elastography: state of the art,'' \emph{European journal of radiology}, vol.~83, no.~3, pp. 429--437, 2014.

\bibitem{cantisani2015strain}
V.~Cantisani \emph{et~al.}, ``Strain us elastography for the characterization of thyroid nodules: advantages and limitation,'' \emph{International journal of endocrinology}, vol. 2015, no.~1, p. 908575, 2015.

\bibitem{onur2015utility}
M.~R. Onur, A.~K. Poyraz, Z.~Bozgeyik, A.~R. Onur, and I.~Orhan, ``Utility of semiquantitative strain elastography for differentiation between benign and malignant solid renal masses,'' \emph{Journal of ultrasound in medicine}, vol.~34, no.~4, pp. 639--647, 2015.

\bibitem{burcher2001deformation}
M.~R. Burcher, L.~Han, and J.~A. Noble, ``Deformation correction in ultrasound images using contact force measurements,'' in \emph{Proceedings IEEE Workshop on Mathematical Methods in Biomedical Image Analysis (MMBIA 2001)}.\hskip 1em plus 0.5em minus 0.4em\relax IEEE, 2001, pp. 63--70.

\bibitem{clements2016evaluation}
L.~W. Clements \emph{et~al.}, ``Evaluation of model-based deformation correction in image-guided liver surgery via tracked intraoperative ultrasound,'' \emph{Journal of Medical Imaging}, vol.~3, no.~1, pp. 015\,003--015\,003, 2016.

\bibitem{virga2018use}
S.~Virga, R.~G{\"o}bl, M.~Baust, N.~Navab, and C.~Hennersperger, ``Use the force: deformation correction in robotic 3d ultrasound,'' \emph{International journal of computer assisted radiology and surgery}, vol.~13, no.~5, pp. 619--627, 2018.

\bibitem{geng2024force}
Y.~Geng \emph{et~al.}, ``Force sensing guided artery-vein segmentation via sequential ultrasound images,'' in \emph{International Conference on Medical Image Computing and Computer-Assisted Intervention}.\hskip 1em plus 0.5em minus 0.4em\relax Springer, 2024, pp. 656--666.

\bibitem{sadigh2012accuracy}
G.~Sadigh, R.~C. Carlos, C.~H. Neal, and B.~A. Dwamena, ``Accuracy of quantitative ultrasound elastography for differentiation of malignant and benign breast abnormalities: a meta-analysis,'' \emph{Breast cancer research and treatment}, vol. 134, pp. 923--931, 2012.

\bibitem{cantisani2014ultrasound}
V.~Cantisani \emph{et~al.}, ``Ultrasound elastography in the evaluation of thyroid pathology. current status,'' \emph{European journal of radiology}, vol.~83, no.~3, pp. 420--428, 2014.

\bibitem{balakrishnan2019voxelmorph}
G.~Balakrishnan, A.~Zhao, M.~R. Sabuncu, J.~Guttag, and A.~V. Dalca, ``Voxelmorph: a learning framework for deformable medical image registration,'' \emph{IEEE transactions on medical imaging}, vol.~38, no.~8, pp. 1788--1800, 2019.

\bibitem{chen2022transmorph}
J.~Chen, E.~C. Frey, Y.~He, W.~P. Segars, Y.~Li, and Y.~Du, ``Transmorph: Transformer for unsupervised medical image registration,'' \emph{Medical image analysis}, vol.~82, p. 102615, 2022.

\bibitem{kim2022diffusemorph}
B.~Kim, I.~Han, and J.~C. Ye, ``Diffusemorph: Unsupervised deformable image registration using diffusion model,'' in \emph{European conference on computer vision}.\hskip 1em plus 0.5em minus 0.4em\relax Springer, 2022, pp. 347--364.

\bibitem{kim2021cyclemorph}
B.~Kim, D.~H. Kim, S.~H. Park, J.~Kim, J.~G. Lee, and J.~C. Ye, ``Cyclemorph: cycle consistent unsupervised deformable image registration,'' \emph{Medical image analysis}, vol.~71, p. 102036, 2021.

\bibitem{zhao2024correspondence}
M.~Zhao, J.~Jiang, L.~Ma, S.~Xin, G.~Meng, and D.-M. Yan, ``Correspondence-free non-rigid point set registration using unsupervised clustering analysis,'' in \emph{Proceedings of the IEEE/CVF Conference on Computer Vision and Pattern Recognition}, 2024, pp. 21\,199--21\,208.

\bibitem{jiang2023robotic}
Z.~Jiang, S.~E. Salcudean, and N.~Navab, ``Robotic ultrasound imaging: State-of-the-art and future perspectives,'' \emph{Medical image analysis}, vol.~89, p. 102878, 2023.

\bibitem{ye2021feasibility}
R.~Ye \emph{et~al.}, ``Feasibility of a 5g-based robot-assisted remote ultrasound system for cardiopulmonary assessment of patients with coronavirus disease 2019,'' \emph{Chest}, vol. 159, no.~1, pp. 270--281, 2021.

\bibitem{suligoj2021robust}
F.~Suligoj, C.~M. Heunis, J.~Sikorski, and S.~Misra, ``Robust--an autonomous robotic ultrasound system for medical imaging,'' \emph{IEEE access}, vol.~9, pp. 67\,456--67\,465, 2021.

\bibitem{jiang2021autonomous}
Z.~Jiang \emph{et~al.}, ``Autonomous robotic screening of tubular structures based only on real-time ultrasound imaging feedback,'' \emph{IEEE Transactions on Industrial Electronics}, vol.~69, no.~7, pp. 7064--7075, 2021.

\bibitem{fu2021biomechanically}
Y.~Fu \emph{et~al.}, ``Biomechanically constrained non-rigid mr-trus prostate registration using deep learning based 3d point cloud matching,'' \emph{Medical image analysis}, vol.~67, p. 101845, 2021.

\bibitem{uzunova2017training}
H.~Uzunova, M.~Wilms, H.~Handels, and J.~Ehrhardt, ``Training cnns for image registration from few samples with model-based data augmentation,'' in \emph{Medical Image Computing and Computer Assisted Intervention- MICCAI 2017: 20th International Conference, Quebec City, QC, Canada, September 11-13, 2017, Proceedings, Part I 20}.\hskip 1em plus 0.5em minus 0.4em\relax Springer, 2017, pp. 223--231.

\bibitem{dosovitskiy2015flownet}
A.~Dosovitskiy \emph{et~al.}, ``Flownet: Learning optical flow with convolutional networks,'' in \emph{Proceedings of the IEEE international conference on computer vision}, 2015, pp. 2758--2766.

\bibitem{zhao2025occlusion}
M.~Zhao, G.~Meng, and D.~Yan, ``Occlusion-aware non-rigid point cloud registration via unsupervised neural deformation correntropy,'' in \emph{The Thirteenth International Conference on Learning Representations}, 2025.

\bibitem{liu2021swin}
Z.~Liu \emph{et~al.}, ``Swin transformer: Hierarchical vision transformer using shifted windows,'' in \emph{Proceedings of the IEEE/CVF international conference on computer vision}, 2021, pp. 10\,012--10\,022.

\bibitem{hu2018weakly}
Y.~Hu \emph{et~al.}, ``Weakly-supervised convolutional neural networks for multimodal image registration,'' \emph{Medical image analysis}, vol.~49, pp. 1--13, 2018.

\bibitem{xu2019deepatlas}
Z.~Xu and M.~Niethammer, ``Deepatlas: Joint semi-supervised learning of image registration and segmentation,'' in \emph{International Conference on Medical Image Computing and Computer-Assisted Intervention}.\hskip 1em plus 0.5em minus 0.4em\relax Springer, 2019, pp. 420--429.

\bibitem{wang2013viscoelastic}
Y.~Wang and M.~F. Insana, ``Viscoelastic properties of rodent mammary tumors using ultrasonic shear-wave imaging,'' \emph{Ultrasonic imaging}, vol.~35, no.~2, pp. 126--145, 2013.

\bibitem{herrmann2018assessment}
E.~Herrmann \emph{et~al.}, ``Assessment of biopsy-proven liver fibrosis by two-dimensional shear wave elastography: An individual patient data-based meta-analysis,'' \emph{Hepatology}, vol.~67, no.~1, pp. 260--272, 2018.

\bibitem{li2022arterial}
G.~Li, Y.~Jiang, Y.~Zheng, W.~Xu, Z.~Zhang, and Y.~Cao, ``Arterial stiffness probed by dynamic ultrasound elastography characterizes waveform of blood pressure,'' \emph{IEEE Transactions on Medical Imaging}, vol.~41, no.~6, pp. 1510--1519, 2022.

\bibitem{menzilcioglu2015strain}
M.~Menzilcioglu \emph{et~al.}, ``Strain wave elastography for evaluation of renal parenchyma in chronic kidney disease,'' \emph{The British journal of radiology}, vol.~88, no. 1050, p. 20140714, 2015.

\bibitem{junker2014real}
D.~Junker \emph{et~al.}, ``Real-time elastography of the prostate,'' \emph{BioMed research international}, vol. 2014, no.~1, p. 180804, 2014.

\bibitem{xu2011eus}
W.~Xu \emph{et~al.}, ``Eus elastography for the differentiation of benign and malignant lymph nodes: a meta-analysis,'' \emph{Gastrointestinal endoscopy}, vol.~74, no.~5, pp. 1001--1009, 2011.

\bibitem{sigrist2017ultrasound}
R.~M. Sigrist, J.~Liau, A.~El~Kaffas, M.~C. Chammas, and J.~K. Willmann, ``Ultrasound elastography: review of techniques and clinical applications,'' \emph{Theranostics}, vol.~7, no.~5, p. 1303, 2017.

\bibitem{ashikuzzaman2019global}
M.~Ashikuzzaman, C.~J. Gauthier, and H.~Rivaz, ``Global ultrasound elastography in spatial and temporal domains,'' \emph{IEEE transactions on ultrasonics, ferroelectrics, and frequency control}, vol.~66, no.~5, pp. 876--887, 2019.

\bibitem{tehrani2022bi}
A.~K. Tehrani, M.~Sharifzadeh, E.~Boctor, and H.~Rivaz, ``Bi-directional semi-supervised training of convolutional neural networks for ultrasound elastography displacement estimation,'' \emph{IEEE Transactions on Ultrasonics, Ferroelectrics, and Frequency Control}, vol.~69, no.~4, pp. 1181--1190, 2022.

\bibitem{jiang2023defcor}
Z.~Jiang, Y.~Zhou, D.~Cao, and N.~Navab, ``Defcor-net: Physics-aware ultrasound deformation correction,'' \emph{Medical Image Analysis}, vol.~90, p. 102923, 2023.

\bibitem{jiang2021deformation}
Z.~Jiang, Y.~Zhou, Y.~Bi, M.~Zhou, T.~Wendler, and N.~Navab, ``Deformation-aware robotic 3d ultrasound,'' \emph{IEEE Robotics and Automation Letters}, vol.~6, no.~4, pp. 7675--7682, 2021.

\bibitem{li2025ultrap}
Y.~Li, K.~W. Kwok, M.~Wysocki, N.~Navab, and Z.~Jiang, ``Ultrap-net: Reverse approximation of tissue properties in ultrasound imaging,'' \emph{Advanced Intelligent Systems}, p. 2400865, 2025.

\bibitem{samei2018real}
G.~Samei \emph{et~al.}, ``Real-time fem-based registration of 3-d to 2.5-d transrectal ultrasound images,'' \emph{IEEE transactions on medical imaging}, vol.~37, no.~8, pp. 1877--1886, 2018.

\bibitem{cao2023ultra}
G.~Cao, M.~Chen, J.~Hu, and H.~Liu, ``An ultra-fast intrinsic contact sensing method for medical instruments with arbitrary shape,'' \emph{IEEE Robotics and Automation Letters}, vol.~8, no.~11, pp. 6955--6962, 2023.

\bibitem{vaswani2017attention}
A.~Vaswani \emph{et~al.}, ``Attention is all you need,'' \emph{Advances in neural information processing systems}, vol.~30, 2017.

\bibitem{avants2014insight}
B.~B. Avants, N.~J. Tustison, M.~Stauffer, G.~Song, B.~Wu, and J.~C. Gee, ``The insight toolkit image registration framework,'' \emph{Frontiers in neuroinformatics}, vol.~8, p.~44, 2014.

\bibitem{avants2011reproducible}
B.~B. Avants, N.~J. Tustison, G.~Song, P.~A. Cook, A.~Klein, and J.~C. Gee, ``A reproducible evaluation of ants similarity metric performance in brain image registration,'' \emph{Neuroimage}, vol.~54, no.~3, pp. 2033--2044, 2011.

\end{thebibliography}
\end{document}